\documentclass[sn-mathphys]{sn-jnl}

\jyear{2021}%

\theoremstyle{thmstyleone}%
\theoremstyle{thmstyletwo}%

\theoremstyle{thmstylethree}%

\usepackage{graphicx}
\usepackage{enumerate}
\usepackage[justification=centering]{caption}
\usepackage{tabularx}
\usepackage{subfigure}
\usepackage[figuresright]{rotating}

\begin{document}

\title[\footnotesize Augmenting Interpretable Knowledge Tracing by Ability Attribute and Attention Mechanism]{Augmenting Interpretable Knowledge Tracing by Ability Attribute and Attention Mechanism}


\author[1]{\fnm{Yue} \sur{Yuqi}}\email{yueyuqi980717@163.com}

\author*[1]{\fnm{Ji} \sur{Weidong}}\email{kingjwd@126.com}
\equalcont{These authors contributed equally to this work.}

\author[1]{\fnm{Sun} \sur{Xiaoqing}}\email{sunxiaoqing2649@163.com}
\equalcont{These authors contributed equally to this work.}

\affil*[1]{\orgdiv{School of Computer Science and Information Engineering}, \orgname{Harbin Normal University}, \orgaddress{\city{Harbin}, \postcode{150025}, \state{HeiLongjiang}, \country{People's Republic of China}}}


\abstract{Knowledge tracing aims to model students' past answer sequences to track the change in their knowledge acquisition during exercise activities and to predict their future learning performance. Most existing approaches ignore the fact that students' abilities are constantly changing or vary between individuals, and lack the interpretability of model predictions. To this end, in this paper, we propose a novel model based on ability attributes and attention mechanism. We first segment the interaction sequences and captures students' ability attributes, then dynamically assign students to groups with similar abilities, and quantify the relevance of the exercises to the skill by calculating the attention weights between the exercises and the skill to enhance the interpretability of the model. We conducted extensive experiments and evaluate real online education datasets. The results confirm that the proposed model is better at predicting performance than five well-known representative knowledge tracing models, and the model prediction results are explained through an inference path.}




\maketitle

\section{Introduction}\label{sec1}

With the continuous development of intelligent education, many open and intelligent teaching platforms such as Intelligent Tutoring System (ITS)\cite{bib1} and Massive Open Online Courses (MOOCs) have been further popularised, and the scale of online learning has been expanding rapidly. As a result, online learning has now become a mainstream way of learning for students\cite{bib2}. In contrast to traditional offline teaching, while online education can fully retain students' learning trajectories\cite{bib3}, the disparity in the number of teachers and students makes it difficult to provide personalized tutoring\cite{bib4}. Therefore, Knowledge Tracing (KT) as a powerful tool for intelligent education, how to use it to study and analyze detailed learning interactions on online platforms to personalize teaching and learning becomes a key issue. From a pedagogical perspective, it is a cyclical process of discovering data from the educational environment and then using that data to improve the educational environment\cite{bib5}.

The knowledge-tracing task can be formalized as a supervised sequential learning task. Students interact with exercises containing different knowledge concepts during the learning process. The knowledge-tracing model aims to model students' knowledge states based on their past interaction records\cite{bib6} to determine how well they understand the knowledge concepts, to continuously track changes in students' knowledge, and thus predict their future answers\cite{bib7}. Among the traditional knowledge tracing models, Bayesian knowledge tracing (BKT) based on Hidden Markov Model\cite{bib8,bib12} has been widely used in intelligent tutoring systems, and subsequent related work has applied Item Response Theory (IRT) models, Learning Factor Analysis (LFA) framework, and Performance Factor Analysis (PFA) framework to knowledge tracing respectively. In recent years, with deep learning models outperforming traditional models in areas such as pattern recognition and natural language processing, a large number of deep knowledge tracing models\cite{bib9} based on deep learning have been widely proposed, achieving significant improvements in their predictive power compared to traditional non-deep models.

In 2015 Piech et al. proposed the seminal Deep Knowledge Tracing (DKT) model\cite{bib9}, which uses a recurrent neural network (RNN) as the basic structure and employs hidden vectors to represent students' knowledge states as a way to predict student performance. However, most current approaches to deep knowledge tracing assume that students' own learning abilities remain the same and that all students have the same abilities during the learning process, ignoring the impact of students' abilities on knowledge levels during the learning process. From a pedagogical concept of view, student learning is an individualized process\cite{bib10}, the learning outcomes and learning strategies of the learning subject are influenced by a variety of elements from both internal and external sources, while the stronger the internal elements of the learner, i.e. his own implicit learning abilities, the more explicit his external learning behavior and the learning outcomes achieved will be\cite{bib11}. Students' learning ability is therefore a core element that influences the level of learning and is an important expression of student subjectivity. In the actual learning process, as students continue to learn, different students receive different improvements in their abilities. Currently, only a few works in the field of knowledge tracing have considered the impact of students' ability on learning. Minn et al. improved the DKT and DKVMN, They reflect students' ability by the percentage of correct answers\cite{bib13,bib26}. However, these studies only consider partial information that reflects learners' abilities and cannot clearly explain the level of mastery of a skill by a student in DKT that the model lacks interpretability. In the online education sector, interpretable predictive results can increase user trust and acceptance of the system.

To address the above problems, this paper proposes a deep knowledge tracing based on the ability attribute (AA-DKT) model that integrates the factors of students' ability attributes for dynamic grouping and combines the attention mechanism to calculate the correlation weights between exercises and knowledge concepts to construct students' changing ability attributes with practice.
 
The main innovations and work of this paper are as follows. First, taking full account of the influence of elements within students, i.e. their ability, on their learning behavior, and constructing ability attribute characteristics based on student response outcomes, response times, and student group averages. Second, considering inter-individual variability of students and changes in ability during the learning process, segmenting students' long-term response records, and calculating ability attribute values for dynamic grouping to make predictions of outcomes more accurate. Last but not least, based on deep neural networks, the DKT is extended by combining attention mechanism to quantify the degree of correlation between exercises and knowledge concepts, enhance model interpretability, and visually explain the reasons for the current prediction results through the inference path between knowledge concepts and exercises.

\section{Related Work}\label{sec2}
The current mainstream knowledge tracing models can be broadly classified into two categories: Knowledge tracing based on the Hidden Markov model and Deep Learning.

\subsection{Bayesian Knowledge Tracing (BKT)}\label{subsec2}
The earlier proposed Bayesian knowledge tracing (BKT)\cite{bib15} is one of the typical models based on the Hidden Markov Model for knowledge tracing.BKT models the change in students' cognitive state by representing students' potential knowledge state of a knowledge concept as a binary variable\cite{bib16}, with variable one learning state as the potential data and variable two answer performance as the observed variable, updating students' knowledge state and hence their performance in future learning according to the Hidden Markov Model. Subsequent scholars have proposed more extensions to the BKT, taking further into account other factors. In 2011, Pardos et al. proposed the KT-IDEM\cite{bib17}, a model that introduced the effect of exercise difficulty on predicting student performance. The study by Baker et al\cite{bib18} introduced the effect of student guesses and errors on predicting student performance. In 2014, Khajah et al. extended human cognitive factors to BKT\cite{bib19}, thereby improving prediction accuracy. However, the Hidden Markov-based knowledge tracing model ignores the correlation between knowledge concepts, and can only model students' mastery of each knowledge concept separately, which is difficult to handle more complex knowledge systems, and only simply records students' mastery of each knowledge concept as "mastered" and "not mastered". It is not possible to model intermediate states and it is difficult to model long interaction sequences in practice.

\subsection{Deep Knowledge Tracing (DKT)}\label{subsec3}
In recent years, with the development of deep learning, more scholars have started to apply deep learning methods to the field of knowledge tracing. Deep Knowledge Tracing (DKT) is the first attempt of recurrent neural networks for the knowledge tracing task, which takes students' historical interaction records as input, represents the hidden state vector of the neural network as students' knowledge level, and predicts students' future learning performance based on this. Compared with BKT, DKT models not only have better predictive performance but also do not require manual annotation of exercises and knowledge concepts, so many scholars focus on this area\cite{bib28}.  In 2018, Chen et al.'s study\cite{bib20} improved the prediction performance of DKT by considering the a priori relationship between knowledge concepts. Su et al. proposed to use the textual information of the exercises and students' learning history as the input of the recurrent neural network\cite{bib21}, and used the hidden vector of RNN to model students' knowledge level, which achieved good prediction results, but still could not model students' mastery of each knowledge concept. In 2016, Zhang et al. proposed the DKVMN model\cite{bib31}, which borrowed the idea of memory networks and used the value matrix to model students' mastery of each knowledge concept, examine the relationship between the exercises and each knowledge concept, and track students' mastery of each knowledge concept. In 2019, Abdelrahman et al improved DKVMN by using the attention mechanism to focus on students' answer history when answering similar exercises\cite{bib22}. Sun et al\cite{bib23} extended the behavioral features of students when answering questions to DKVMN to achieve better prediction results. In 2019, Liu et al proposed EKT \cite{bib24} to use the textual information of the exercises with students' learning history for knowledge tracing. Nakagawa et al. proposed GKT \cite{bib25} to model the relationship between individual knowledge concepts with a knowledge graph, achieving good prediction results. In 2019, Cheng et al. proposed DIRT\cite{bib27} to extend the traditional method item response theory through deep learning. Ai et al.'s work\cite{bib29} considered the containment relationship between knowledge concepts and re-designed the memory matrix and improved DKVMN with good results.

\subsection{Interpretable works}\label{subsec2}
Deep learning uses an end-to-end learning process, and the model training process and final results often lack interpretability; in the DKT task, the student's hidden knowledge state is an extremely abstract vector that does not possess interpretability\cite{bib14}. To achieve an interpretable knowledge tracing model, in 2019 Liu et al. proposed EKT\cite{bib24} to calculate the similarity between exercises and use the correlation weights as weights of knowledge states to enhance model interpretability, in 2020 Liu et al. proposed A-DKT\cite{bib30} to calculate the attention weights between knowledge concepts using Jaccard coefficients, in 2020 Ghosh proposed AKT\cite{bib32} model, which applies contextual information combined with IRT model to enhance the representational power of embedding and provide interpretation for the results.

In general, the current studies have been able to track students' mastery of various knowledge concepts to some extent, but most have neglected the influence of students' own abilities on the learning process, or have not modeled students' abilities comprehensively enough and lacked interpretability. That is, the default is that students have the same level of ability before and after the learning interval and that there is no differentiation of ability between individuals, yet in the actual learning process, students' learning ability tends to change constantly according to their previous learning. In response to this question, in this paper, we propose a deep knowledge tracing model that integrates ability attributes and attention mechanism, integrated taking into account the factors that affect students' own ability and using students' exercise results as indirect feedback on their knowledge mastery. Modeling changes in students' abilities, tracking their mastery of knowledge, and predicting their future performance.

\section{Problem Definition}\label{sec3}
In this paper, we define the set of students as $ |S| $, the set of exercises $ |E| $ , and the set of knowledge concepts as $ |K| $. Suppose that in a learning scenario, each student independently answers the exercises, we record the answer history of a certain student as $ X_{t}=\left[\left(e_{1}, r_{1}\right),\left(e_{2}, r_{2}\right), \ldots,\left(e_{t}, r_{t}\right)\right] $, where $ e_{t} \in E $, is the exercise answered by the student at the time  $ T $; $ r_{t} $  denotes the corresponding answer result, and if the student answers exercise right, then $ r_{t} $ equals to 1, otherwise =0 if the answer is incorrect. The matrix $ M^{K}\left(d_{k} \times|K|\right) $ is defined as the embedding representation of $ K $ knowledge concepts in the entire knowledge space, and the $ d_{k} $ dimensional column vector is the embedding representation of a knowledge concept. Since the evaluation of the knowledge state cannot be quantified externally, the existing knowledge tracing task predicts the probability of a user answering an exercise correctly at the next moment as $ y_{t+1}=P\left(r_{t+1}=1 \mid e_{t+1}, X_{t}\right) $ .

Based on the above description, we can define the problem as:

The achievement of the following objectives is based on the learning history of each student:(1) Tracing changes in the state of pupils' knowledge;(2) Predict the outcome of the student's answer $ r_{t+1} $ on the next exercise $ e_{t+1} $ . 

\section{Proposed Approach}\label{sec4}
Students' learning ability is one of the elements that affect the level of learning. In the actual learning process, as students continue to learn, their own ability grows and varies greatly from student to student. Based on this idea, this paper proposes a knowledge tracing model (AA-DKTA) that integrates ability attribute and attention mechanism. The model segmenting students' long-term answer interaction sequences, constructing ability attributes based on students' answer results and response times, obtains the ability changes of students during their learning process by and then dynamically assigning students to similar abilities by K-means. Finally, use the deep knowledge tracing model incorporating attention mechanism to quantify the correlation between exercises and knowledge concepts to model student learning behaviour and predict student performance at the next moment.

Along this line, Fig. 1 shows the overall structure of the AA-DKTA model.

\begin{figure}[h]
	\centering
	\includegraphics[width=0.75\textwidth]{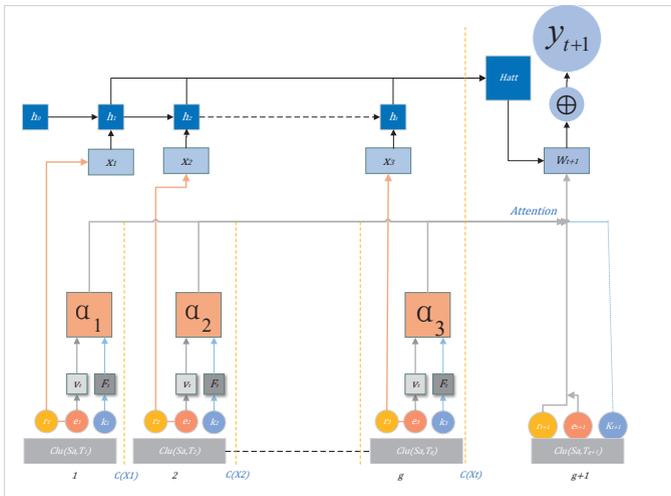}
	\caption{The structure of the AA-DKTA model}\label{fig1}
\end{figure}

\subsection{Ability attribute}\label{subsec2}
Segmenting students into groups of similar ability based on historical ability performance can be more effective in providing more targeted instruction. Students' ability attributes change dynamically over time, so we use five time steps as unit lengths to divide students' interaction records over time, as shown in Fig. 2.

\begin{figure}[h]
	\centering
	\includegraphics[width=0.75\textwidth]{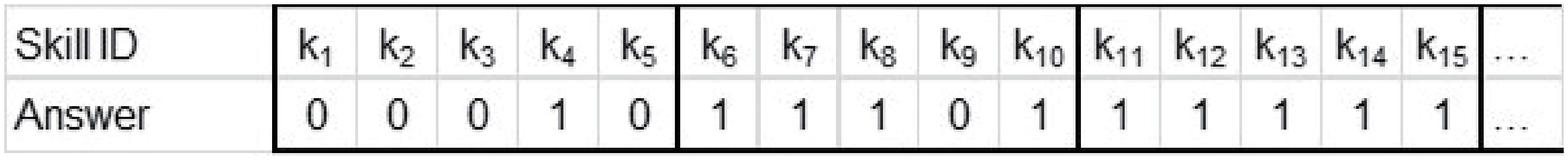}
	\caption{Student interaction record division}\label{fig2}
\end{figure}

Fig. 2 shows a sequence of 15 attempted responses from a student, divide it into three time segments, where each segment represents the time interval between the student's responses to the five questions in the system. The learning recorded for each segment reflects the student's ability to learn over that period of time, and the student's ability is reassessed after each time segment and is continuously and students dynamically assigned to the group to which the next segment belongs. At the same time, the division of student answer sequences can somewhat reduce the allocation of memory space for long sequences of learning.

The value of a student's ability attribute reflects to some extent the student's mastery of the knowledge he has acquired. Since a student's answer record is a time series of data, changes in the student's learning ability can be reflected according to the student's speed in doing the question and the change in the result, and we construct the student's ability attribute with this in mind. For ease of calculation, the ratio of the student's response time to correctly answer the question to the average response time of all students is defined as the student's ability attribute, as shown in Eq.(1) and Eq.(2):

\begin{equation}
	C\left(e_{j}\right)_{1: g}=\left\{\begin{array}{l}
		\frac{t_{j}}{A_{i j}}, e_{j}==1 \\
		0, e_{j}==0
	\end{array}\right.
\end{equation}

\begin{equation}
	A_{i j}=\sum_{j=1}^{g} \frac{t_{i j}^{s k i l l}}{M}
\end{equation}

where $ C\left(e_{j}\right)_{1 \mathrm{~g}} $ denotes the change in student ability for the question $ j $ over a sequence interval of length $ g $ questions. $ t_{j} $ denotes the response time for a student to correctly answer the exercise $ j $, $ A_{ij} $ denotes the average response time of all students who correctly answered question $ j $ containing knowledge concept $ i $. $ e_{j}==1 $ means the student answered correctly, $ e_{j}==0 $ means the student answered incorrectly , knowledge concept $ i $ contains $ M $ records of correct responses, where the $ j $ response time is $ t_{i j}^{s k i l l} $.
The values of students' competence attributes at the knowledge concept $ i $ were transformed into vectors for subsequent clustering at time interval $ g $, as shown in Eq. (3).

\begin{equation}
	r_{1: g}^{i}=\left(C\left(x_{1}\right)_{1: g}, C\left(x_{2}\right)_{1: g}, \ldots, C\left(x_{n}\right)_{1: g}\right)
\end{equation}

$ r_{1 :  g}^{i} \in R $denotes an ability vector in the time interval $ 1:g $, $ i $ is the number of knowledge concepts, initialised with an $ i $ dimensional all-zero vector, i.e. $ r_{1: g}^{i} \leftarrow[0, \ldots, 0] $ ; when the student answers question $ j $ correctly, the ability on the knowledge concept $ i $ for the student will be boosted by $ C\left(x_{j}\right)_{1: g} $ .

\subsection{Clustering grouping}\label{subsec2}
In this paper, students are divided according to their learning ability by K-means clustering, i.e. students are dynamically assigned to groups with similar learning ability according to their learning ability on each division interval. Before clustering, the number of clusters was first specified. After experimental comparison tests with different values of $ K $ (see 5.3 for the specific experimental results), the number of clusters was set to $K=10$, i.e. students were clustered into 10 groups with different learning abilities.
In the clustering training process, the centre of each student group is first found without considering the time division interval, and after the centre is determined, it is no longer changed throughout the clustering process, and then students are assigned to different groups at each time interval $ T_{g} $. The clustering grouping process is as in Eq. (4).

\begin{equation}
	\operatorname{Clu}\left(S_{a}, T_{g}\right)=\operatorname{argmin} \sum_{k=1}^{|K|} \sum_{1_{1 : g-1}^{i} \in K}\left\|r_{1: g-1}^{i}-\mu_{k}\right\|^{2}
\end{equation}

where $ S_{a} $  denotes student $ a $, $ T_{g} $ denotes a time interval, $\mu _{k}$ represents the centre of a student subgroup $K$, and $ r_{1: g-1}^{i} $ denotes the student's ability attributes over the time series $ 1: g-1 $.

\subsection{Deep Knowledge Tracing on Attention Mechanism (DKTA)}\label{subsec2}
In the DKT model, the hidden state is used to represent the student's mastery of the knowledge concept, however this hidden state is not interpretable and therefore the DKT model is unable to output the student's knowledge level in detail. This paper incorporates an attention mechanism based on deep knowledge tracing, which is reflected in the calculation of the correlation weights between the exercises and the knowledge concepts.

First, the input student's current exercise question $ e_{t} $ is multiplied with the exercise embedding matrix $ B\left(d_{k} *|E|\right) $ to obtain the embedding vector $ v_{t} $ of dimension $ d_{k} $. Let $ x_{t} $ denote the one-hot encoding of the exercise record $ \left(e_{t}, r_{t}\right), x_{t} \in R^{2 m} $, and for the exercise record, RNN (recurrent neural networks) is used to encoding to obtain the hidden layer states, as shown in Eq (5).

\begin{equation}
	h_{t}=R N N\left(h_{t-1}, x_{t-1}\right)
\end{equation}

Mapping the topics to a continuous vector space, building a matrix $ F_{t}\left(|E|^{*} d_{k}\right) $ to store the knowledge concepts covered by the exercises, each $ d_{k} $-dimensional column vector is the embedding vector of a knowledge concept covered by the exercise, calculating the inner product between $ v_{t} $ and knowledge concepts $ F_{t}(i) $, then calculating its $ softmax $  value and stored in the vector $\alpha_{t}(i)$, the knowledge-related weights are shown in Eq (6).

\begin{equation}
	\alpha_{t}(i)=\operatorname{softmax}\left(v_{t}^{T} F_{t}(i)\right)
\end{equation}

Based on the attention weights, the student's mastery of the topic $ w_{t} $ is calculated as shown in Eq (7).

\begin{equation}
	w_{t}=\sum_{i=1}^{N} \alpha_{t}(i) F_{t}^{v}(i)
\end{equation}

The embedding vector $ e_{t} $ of the question implicitly contains difficulty information, which combined with $ w_{t} $ yields $ p_{t} $ containing difficulty and knowledge mastery, then combining deep knowledge tracing model with attention value to predictive results $ y_{t} $ , as in Eq. (8) and Eq. (9).

\begin{equation}
	p_{t}=\tanh \left(W 1\left[w_{t} \oplus e_{t}\right]+b_{1}\right)
\end{equation}

\begin{equation}
	y_{t}=sigmoid\left(W_{2} p_{t}+b_{2}\right)
\end{equation}

where $ \oplus  $ is the tandem operation and $ \left\{W_{1}, W_{2}, b_{1}, b_{2}\right\} $ are the parameters. In the model, the DKTA framework represents the student's state of knowledge acquisition by means of weights $ \alpha_{t}(i) $ for the purpose of improving the interpretability of the model.

\subsection{Interpretability analysis of prediction results}\label{subsec2}
This section illustrates the interpretability of the prediction results through inference paths generated based on the weights associated between the exercises and the knowledge concepts. Fig. 3 shows the record of a student $ s $'s answer on knowledge concept $ k $, predict the process of generating inference paths for the predicted outcome in case this learner answers question $ e_{5} $ incorrectly. Where $ e_{1}, e_{2}, e_{3}, e_{4} $ are the exercises related to knowledge concept $ k $, $ e_{5} $ is the exercise to be predicted, and the correlation weight between $ e_{5} $ and knowledge concept $ k $ is $ \alpha_{2} $. Taking the inference path shown in the figure as an example, we add knowledge concept $ k $ and exercise $ e_{4} $ to the inference sequence. The AA-DKTA generates an inference path for the user's answer to the current exercise: $ e_{5}-k-e_{4} $. Based on the significance of the correlation weight $ \alpha $ between the exercise and the knowledge concept, the model predicts that the reason for the user's incorrect answer to exercise $ e_{5} $ is that knowledge concept $ k $ was examined in past interactions, and knowledge concept $ k $ is correlated with question $ e_{4} $, thus leading to the interpretation that since the learner has incorrectly answered question $ e_{4} $ and therefore has a low level of mastery of knowledge concept $ k $, resulting in incorrect answers to exercise $ e_5 $.

\begin{figure}[h]
	\centering
	\includegraphics[width=0.75\textwidth]{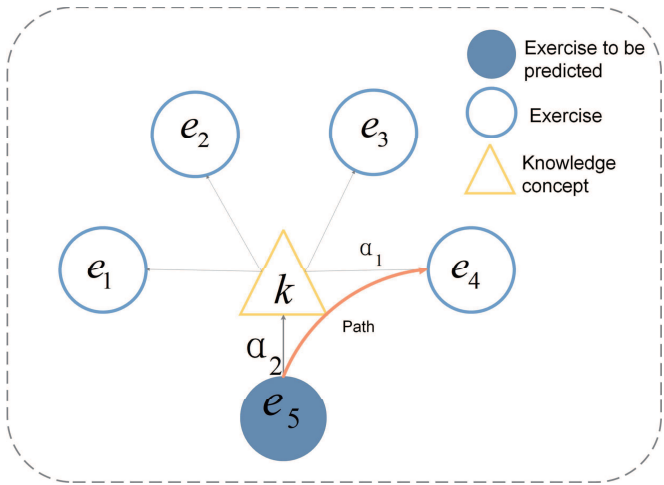}
	\caption{Interpretable reasoning path generation}\label{fig3}
\end{figure}

\subsection{Optimization}\label{subsec2}
In this paper, stochastic gradient descent is used to optimise the model parameters by minimising the cross-entropy loss function between the predicted probabilities and the labels. The loss function is shown in Eq (10).

\begin{equation}
	L=-\sum_{t}\left(r_{t} \log p_{t}+\left(1-r_{t}\right) \log \left(1-p_{t}\right)\right)
\end{equation}

Where $ r_t $ is the actual student response result, where correct is 1 and vice versa, and  $p_t$ is the probability of predicting a correct user response, $ p_{t} \in[0,1]  $.

\section{Experiments}\label{sec5}
To verify the validity of the proposed model in this paper, this section compares it with five other knowledge tracing models on three online education datasets from two different teaching and learning scenarios, followed by ablation experiments to verify the effectiveness of the two modules of fused ability attributes and attention mechanism.

\subsection{Datasets}\label{subsec2}
Three public education datasets were used for the experiments in this paper: ASSISTments2009\cite{bib34}, ASSISTments2017\cite{bib34} and KDDCup2010\cite{bib33}.

KDDCup2010: Developed for the 2010 KDD Cup Educational Data Mining Challenge, this dataset was designed to predict student performance on mathematics problems and contains 607,026 interactions from 574 students on 436 knowledge concepts.

ASSISTment2009: This dataset, from the ASSISTments online learning platform, contains high school mathematics exercises and is a classic dataset in the knowledge tracing domain. 4151 students with 325637 interactions on 110 knowledge concepts.

ASSISTment2017: This dataset is the most recent dataset from ASSISTmentsData and contains 942,816 interactions from 686 students on 102 knowledge concepts.

\subsection{Evaluation indicators}\label{subsec2}
In line with most work in the field of knowledge tracing, this experiment uses the area under the curve (AUC) of the receiver operating characteristic (ROC) curve to measure model performance \cite{bib31}, which has good performance measurement properties. When the AUC is 0.5, the value is the prediction performance obtained by random guessing, and the closer the AUC value is to 1.0, the better the model prediction is. For each model, 20 tests are conducted in this paper and the average AUC value is taken.

\subsection{Experimental setup}\label{subsec2}
The model was implemented using the Pytorch framework with a K-fold cross-validation (K=5) randomly selected and partitioned dataset. In each experiment, 60$\%$ of the students were used as the training set, 20$ \% $ as the test set and 20$ \% $ as the validation set, and the model was evaluated on the cross-validation set after training each epoch. The learner interaction sequence length in this experiment was set to 200 and the batch size was 24. Since too small a K-means clustering number would affect the effectiveness of the experiment and too large a number would increase the computational complexity, a value of 10 was determined based on the comparative test results under different K values, and the test results are shown in Table 1.

\begin{table}[h]
	\caption{Comparison of AUC values at different hyperparameter settings}\label{tab1}
	\newcommand{\tabincell}[2]{\begin{tabular}{@{}#1@{}}#2\end{tabular}}
	\begin{tabular}{c c c c c}
		\hline
		\tabincell{c}{Model} & ASSISTments2009 & ~ & ASSISTments2017  & ~  \\
		 \hline
		\multirow{4}{*}{AA-DKTA} & K & AUC & K & AUC  \\ 
		
		 & 3 & 0.786 & 3 & 0.686 \\ 
		 
		 & 5 & 0.799 & 5 & 0.692  \\ 
		 
		 & 10 & 0.816 & 10 & 0.718   \\ 
		 \hline
\end{tabular}
\centering
\end{table}

\subsection{Analysis of predicted student performance results}\label{subsec2}
To evaluate the performance of the AA-DKTA model, in this paper, five typical benchmark models are selected, namely the two classical models of Bayesian Knowledge Tracing (BKT), Deep Knowledge Tracing (DKT), and the AKT, SAKT and DKT-DSC models based on DKT.

Among them, BKT is one of the classical traditional models, which manually defines the factors affecting student performance through a non-deep learning approach and fits parameters based on user interaction sequences; DKT is a classical deep learning-based model, which learns interaction sequences using RNN modelling and is a pioneering work in the field of deep knowledge tracing; there are many extended models in the field of deep knowledge tracing, and this paper selects three models that have commonalities with The AKT model introduces contextual information and combines it with the IRT model to enhance the representational power of embeddings and provide explanations for the results through an attention mechanism; SAKT is a knowledge tracing model based on the Transformer model and can be seen as a special case of AKT; the DKT-DSC model represents students' ability through the correctness of exercises applied to the classical DKT model.

The experimental results are shown in Table 2, where the bolded data indicate the best results in the data set.

\begin{table}[h]
	\caption{Model performance comparison experimental results (AUC)}\label{tab2}
	\newcommand{\tabincell}[2]{\begin{tabular}{@{}#1@{}}#2\end{tabular}}
	\begin{tabular}{c c c c c c c}
		\hline
		\tabincell{c}{Dataset} & BKT & DKT & SAKT & AKT & DKT-DSC & AA-DKTA \\ \hline
		ASSISTments2009 & \~0.69 & 0.768 & 0.752 & 0.792 & 0.777 & 0.816 \\ \hline
		ASSISTments2017 & -- & 0.721 & 0.6569 & 0.756 & -- & 0.768 \\ \hline
		KDDCup2010 & -- & -- & -- & -- & 0.723 & 0.736 \\ \hline
	\end{tabular}
\centering
\end{table}

The experimental data for DKT, AKT and DKT-DSC were obtained by re-implementing the original method, while the experimental data for BKT and SAKT were obtained from the work\cite{bib31} and\cite{bib32} respectively.

Analysis of the data in the table reveals:

(a) An extended model incorporating ability attribute features for modelling inter-individual differences can further improve the predictive power of deep knowledge tracing models, demonstrating that fully introducing rich feature information from the educational environment and taking into account individualisation is an effective means of improving model performance;

(b) Compared to the DKT-DSC model, the AA-DKTA, which considers student response time and answer outcomes for each knowledge concept and incorporates attention mechanism, is more effective, with the AUC index improving by 4.78$\%$ and 2.94$\%$ on ASSISTments2009 and KDDCup2010, respectively.

According to the experimental results, the deep learning-based methods generally outperformed BKT. DKT used only recurrent neural networks, used hidden vectors to represent students' overall knowledge level and did not take into account students' own ability differences, so DKT's prediction performance was lower than AKT and DKT-DSC, and AA-DKTA model’s average AUC values on all three datasets were better than the other five comparison methods.

The main part of this paper is the construction of the ability attributes and the attention mechanism. The ability attributes can improve the model performance to some extent, and we believe that the attention mechanism can not only provide an explanation for the prediction results, but also improve the prediction ability to some extent. To verify the influence of the two modules in AA-DKTA, an ablation experiment is conducted in this paper. In particular, model AA-DKT incorporated only student ability and model DKT-A incorporated only the attention mechanism. The results of the experiments are shown in Table 3. Both models AA-DKT and DKT-A show significant improvements compared to the DKT model, indicating that the complete personalised learning mechanism can better fit the real student learning situation.

\begin{table}[h]
	\caption{Comparative results of ablation experiments (AUC)}\label{tab3}
	\newcommand{\tabincell}[2]{\begin{tabular}{@{}#1@{}}#2\end{tabular}}
	\begin{tabular}{c c c c c}
		\hline
		\tabincell{c}{Dataset}& DKT & DKT-A & AA-DKT & AA-DKTA \\ \hline
		ASSISTments2009 & 0.768 & 0.772 & 0.797 & 0.816 \\ \hline
		ASSISTments2017 & 0.721 & 0.736 & 0.750 & 0.768 \\ \hline
	\end{tabular}
\centering
\end{table}

\subsection{Tracking the state of knowledge}\label{subsec2}
The objective of the knowledge tracing task is also to track changes in students' knowledge. In this paper, we track students' mastery of the knowledge concepts by capturing a section of a learner's exercise record from the ASSISTments2009 dataset, which contains the student's interactions for 21 exercises on four knowledge concepts, as shown in Fig.4. The horizontal axis of the graph shows the sequence of student responses and the vertical axis shows the knowledge points corresponding to the questions answered during that period of time.

\begin{figure}[h]
	\centering
	\includegraphics[width=.75\textwidth]{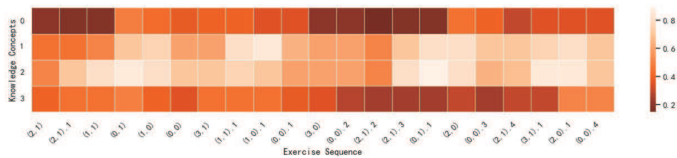}
	\caption{ASSISTments2009 dataset knowledge level output results}\label{fig4}
\end{figure}

According to the results of the experiment, taking knowledge concept 2 as an example, we can see that:

1) At moments 1 and 2, the tracing results of the model's mastery of knowledge concept 2 increased significantly (output values became higher and colour blocks became lighter) when students answered the exercises correctly in succession; while at moment 16, the tracing results decreased after students answered the exercises corresponding to knowledge concept 2 incorrectly, indicating that the model was able to track students' mastery of knowledge concepts based on their performance extent and modelling the learning process.

(2) Observing student’s records from moment 3 to moment 12, student did not answer the exercises corresponding to knowledge concept 2, but his knowledge state kept changing with the interactive sequence of other knowledge concepts. It can be seen that the model proposed in this paper is able to take into account to a certain extent the individuality of the student, meets the average person's intuitive sense of knowledge mastery, and tracks the student's mastery of each knowledge concept in real-time.

Fig.5 shows a comparison of a student's ability attributes before and after practising 24 exercises in the education scenario. The radar chart gives a visual impression of the improvement in their ability level after practising.

\begin{figure}[h]
	\centering
	\includegraphics[width=0.4\linewidth]{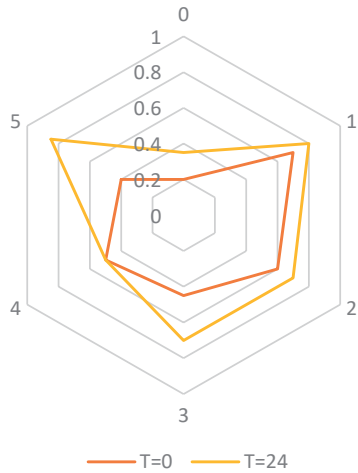}
	\caption{Ability attribute values before and after learning}\label{fig5}
\end{figure}

In this paper, a student's learning interaction records over 8 time steps were extracted from the ASSISTments2009 dataset and their levels of mastery of two knowledge concepts were output using the DKT model and the AA-DKTA model respectively. The student's answer sequences were:

$ H_{s} $ = [(0,1),(0,1),(1,1),(0,1),(1,0),(1,1),(0,0),(1,0),(1,0)].

Each item in the sequence indicates the results of the student's answers, and the tracing of knowledge mastery by the 2 groups of models is shown in Fig.6.


\begin{figure}[H]
	\centering  
	\subfigcapskip=-5pt 
	\subfigure[KC1]{
		\includegraphics[width=0.48\linewidth]{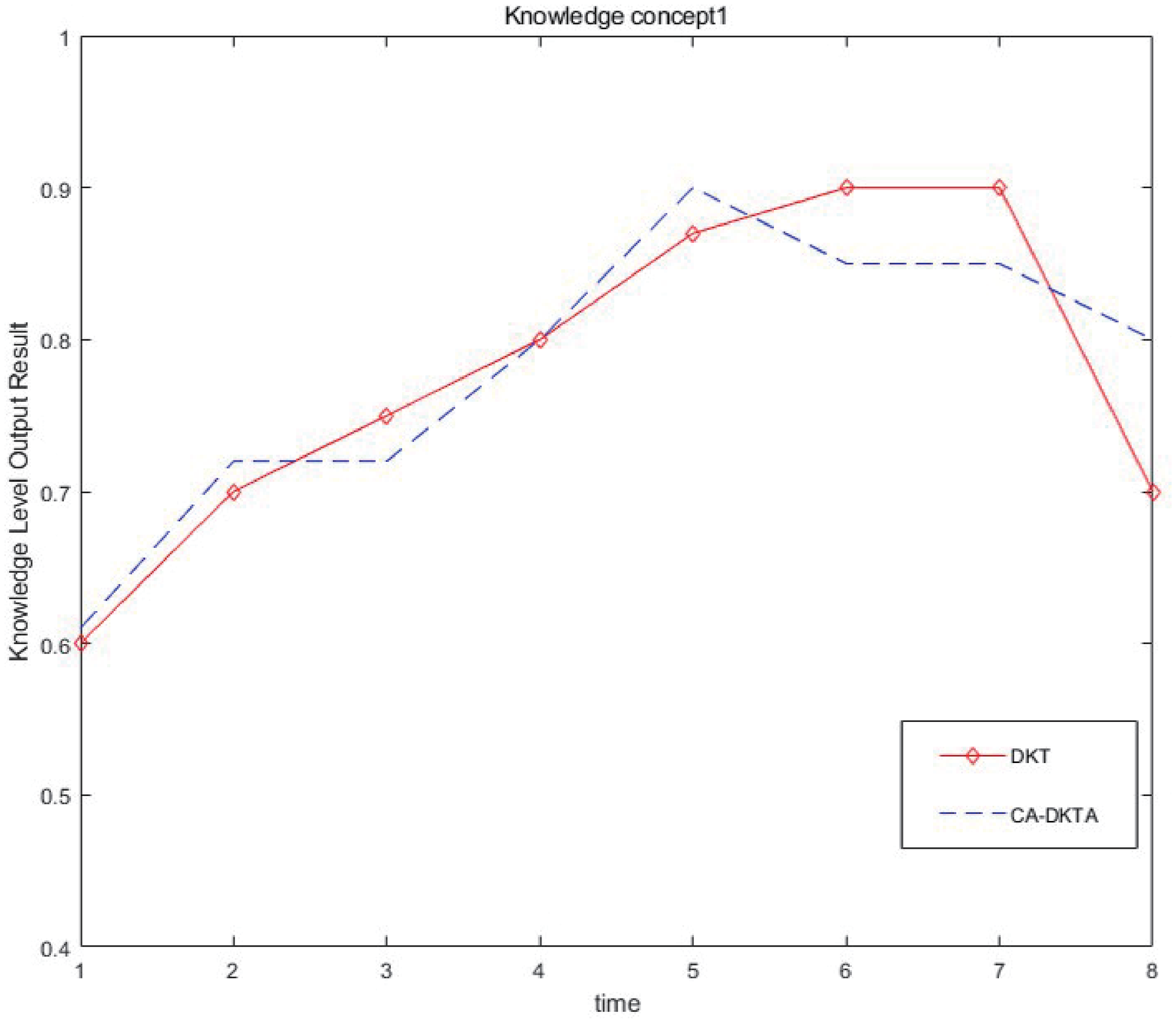}}
	\subfigure[KC2]{
		\includegraphics[width=0.48\linewidth,height=0.4\linewidth]{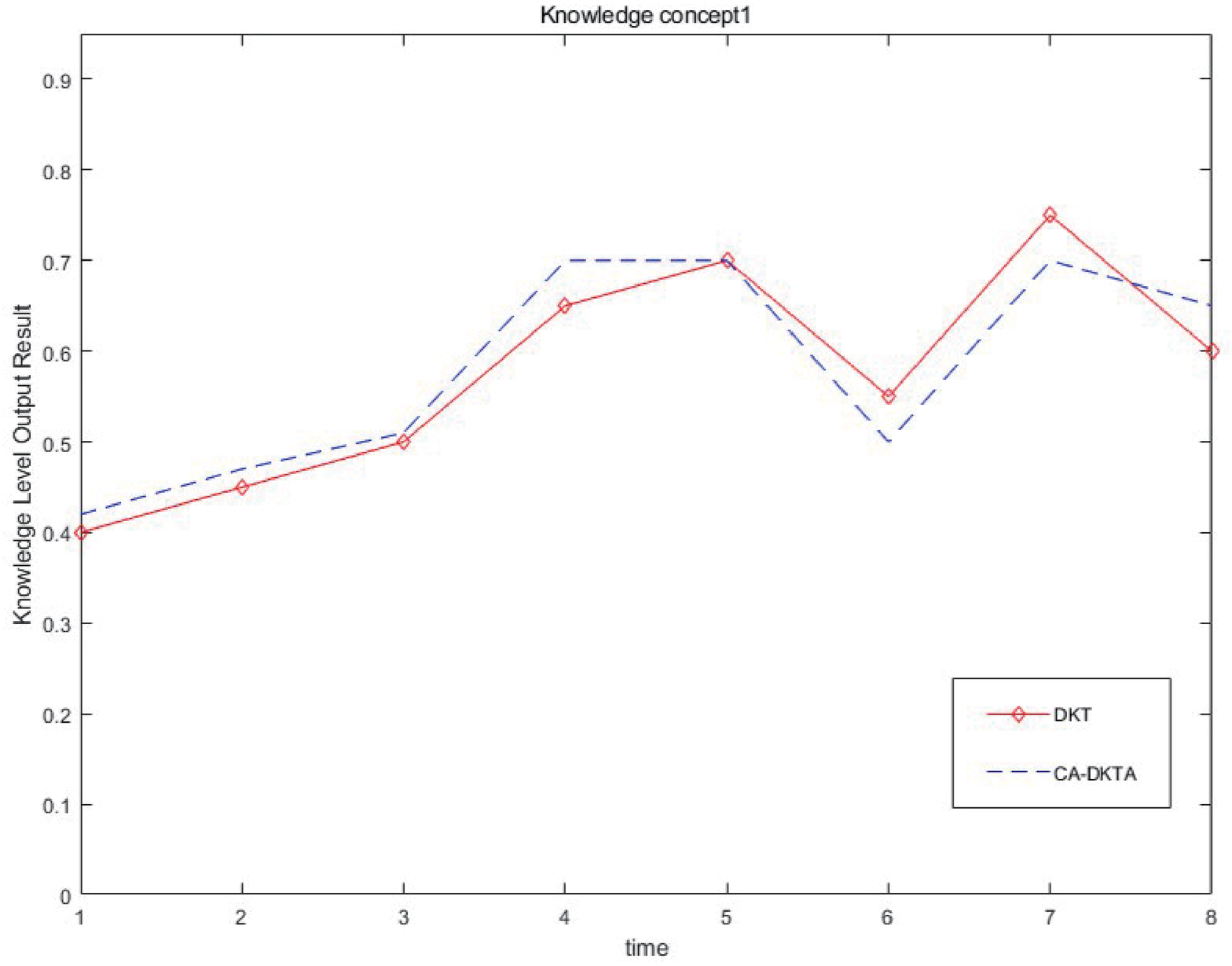}}
	\caption{Comparison of knowledge level output}\label{6}
\end{figure}

The blue dashed line indicates the tracing of the AA-DKTA model for the student's knowledge level, and the red line indicates the tracing of the DKT model. In Fig. 6(b), the student correctly answers the exercise containing knowledge concept 2 at moment 3, and both sets of models reflect an increase in knowledge level at moment 3. It explains that both DKT and AA-DKTA can be used to model learning behavior and track knowledge levels. The DKT output showed relatively less change in knowledge level after the student incorrectly answered the question at moment 5, but the AA-DKTA showed a continuous decrease, indicating that the AA-DKTA takes into account other relevant factors of the student and constructs the student's ability attributes to track the effect of changes in the student's ability on knowledge acquisition.

\section{Conclusion}\label{sec4}

This paper considers the influence of students' individual ability factors on their learning process, and proposes a deep knowledge tracing model AA-DKTA, which integrates ability attributes and attention mechanism, and takes into account the changing abilities of students in the learning process and the differences between individuals, dynamic grouping is carried out and combined with attention mechanism, calculation of correlation weights between exercises and knowledge points to further explain the students' state of mastery of knowledge. The experimental results show that the AUC of AA-DKTA is significantly higher than the other five methods on the three open data sets, which validates the effectiveness of the proposed method in this paper. In subsequent research, we will address the following areas:

(1) Online education platforms provide a large amount of students' learning information and trajectories, how to better extract and model these features to achieve more accurate predictions needs further consideration. 

(2) The real-world learning environment is complex, and the next step is to use methods from fields such as natural language processing to deal with heterogeneous data, such as textual information on topics and multiple types of interaction information, and how to better process knowledge points and topics for comprehensive topics with high knowledge coverage.

\bibliography{KT.bib}

\newpage

\section*{Declarations}

\noindent \textbf{Authors}: 

1. Yue Yuqi, Female, born in 1998, master's degree, research direction is knowledge tracing,swarm intelligence.

2. Ji Weidong, Male, born in 1978, Doctor, Professor, research direction is swarm intelligence,big data.

3. Sun Xiaoqing, Female, born in 1994, master's degree, research direction is knowledge tracing, swarm intelligence.

\noindent \textbf{Data Access Statements}:

This publication is supported by multiple datasets, which are openly available at locations cited in the reference section.

\noindent \textbf{Editorial Board Members and Editors}:

The author has no competitive interest with members of the Editorial Committee and Editors.

\noindent \textbf{Funding}:

1. National Natural Science Foundation of China(31971015)

2. Natural Science Foundation of Heilongjiang Province in 2021(LH2021F037)

3. Special research project on scientific and technological innovation talents of Harbin Science and Technology Bureau(2017RAQXJ050)

4. Harbin Normal University Graduate Student Innovation Project (HSDSSCX2022-16)

\noindent \textbf{Competing financial interests}: 

The authors have no competing interests to declare that are relevant to the content of this article.

\newpage

\end{document}